\documentclass[journal]{IEEEtran}
\usepackage{graphicx}
\usepackage{amsmath}
\usepackage{url}
\usepackage{hyperref}
\ifCLASSOPTIONcompsoc
	\usepackage[caption = false, font = normalsize, labelfont = sf, textfont = sf]{subfig}
\else
	\usepackage[caption = false, font = footnotesize]{subfig}
\fi

\title{Designing an Inertia Actuator with a Fast Rotating Gyro inside an Egg-shaped Robot}

\author{Chun-Chi~Wang, He-Zhi~Liu, Rui-Yuan~Lin, Li-Yang~Lu and N. Michael~Mayer  
	\thanks{Research supported by Advanced Institute of Manufacturing with High-tech Innovations (AIM-HI)}%
	\thanks{Chun-Chi~Wang and He-Zhi~Liu was with Department of Electrical Engineering, National Chung Cheng University, Chiayi, 621, Taiwan. (e-mail: s8704112002@hotmail.com, hayden0828@gmail.com).}%
	\thanks{Rui-Yuan~Lin and Li-Yang~Lu is with Department of Electrical Engineering, National Chung Cheng University, Chiayi, 621, Taiwan. (e-mail: e4522121@gmail.com, leon157375@gmail.com).}%
	\thanks{N.Michael~Mayer is with the Department of Electrical Engineering, National Chung Cheng University, Chiayi, 621, Taiwan (corresponding author, e-mail: nmmayer@gmail.com).}%
}

\begin{document}
	\maketitle
	
	
	
	\begin{abstract}\label{abs:Abstract}
In this paper, we describe features of two new robot prototypes that are actuated by an actively controlled gyro (flywheel, symmetric rotor) inside a hollow sphere that is located in the middle of the robots. No external actuators are used. The outside structure of the robots and the gyro are connected by a gimbal, which is similar in structure to a control moment gyroscope in spacecrafts. The joints of the gimbal can be actuated. In this way, the orientation of axis for the gyro in relation to the egg can be changed. Since the inertia of the fast rotating gyro is large in relation to the outside structure, a relative rotation of the axis against the outside structure results in a motion of the egg by inertia principle. In this way, we can use this principle for controlling the robot to move forward and turn around.
The robots are shaped as spheroidal ellipsoids so they resemble eggs.   So far, we have built and tested two robot prototypes.
	\end{abstract}


	\section{Introduction and related research}\label{sec:introduction and related research}

\IEEEPARstart{E}{arlier} projects of gyro actuated robots are the Gyrover at the Carnegie Mellon University \cite{tsai1999control} and Gyrobot projects \cite{1569185}. They use a fast spinning gyro mounted in a wheel-shaped robot, balancing the robot by tilting the flywheel. And forward motion is produced by another motor.
		
Joydeep Biswas \cite{reaction_wheel_robot} implements the wheel robot Reactobot by actuating an internal reaction wheel. The axis of the reaction wheel is the same as the forward direction, turning the reaction wheel can affect the robot turning in motion.
		
Gyros have been used for the stabilization of biped robots \cite{doi:10.1533/abbi.2006.0032}\cite{4058571}. Further ideas were to stabilize also the pitch of the robot by using the gyro at the same time as a reaction wheel \cite{jumpingjoe1}.
		
The idea came up to allow the robot to make rapid movements by adding a brake mechanism (same as a car brake) to produce a large torque in order to allow rapid movements such as somersault and quickly standing up.
		
		\IEEEpubidadjcol
		
Jumping Joe robots were presented at the NEDO hall at the AICHI world exhibition (EXPO 2005), near Nagoya, Japan. The momentum is produced by stopping the rotor with a car brake. In that way, a very strong momentum has been achieved and the robot could stand up rapidly. Similar principles have been used in the Cubli robot
\cite{cubli}.
		
In spacecraft or satellite, a momentum wheel (or gyroscope) is used for gyroscopic stabilization and orientation. The principle is conservation of angular momentum- accelerating a reaction wheel brings about a proportional response by the rest of a spacecraft or satellite. In addition, designs similar to those outlined here for our robot are known there as Controlled Momentum Gyroscopes \cite{bailey2000orienting}.

\begin{figure}[b]
\centerline{\includegraphics[width=0.45\textwidth]{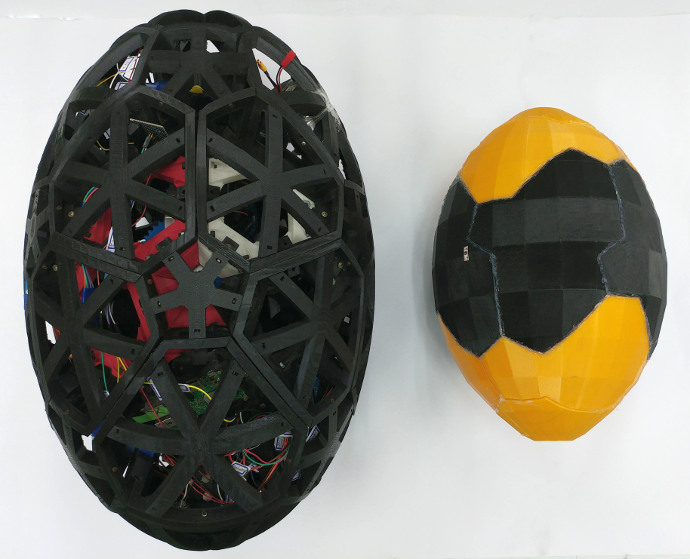}}
\caption{
\label{fig:robo_foto} Our two robot prototypes. 1st prototype (left, 63x40cm), 2nd prototype (right, bumblebee,  44x32cm).
}
\end{figure}

Recently, more approaches have been brought up in robotics. One example is bicycle robots manufactured by Murata (as Murata boy on their webpage) \cite{murata} and other companies. In addition, Northwestern University recently presented a Conservation of Angular Momentum Locomotion Robot (Fluffbot) on their webpage \cite{fluffbot}.
		
Various toys are based on gyro effects. One example is the Dynabee device that can be used to train the hand muscles that use outside applied torque for further acceleration. The physics of the Dynabee is outlined in \cite{dynabee}.

\IEEEpubidadjcol

\begin{figure*}[!t]
\centerline{\includegraphics[width=0.89\textwidth]{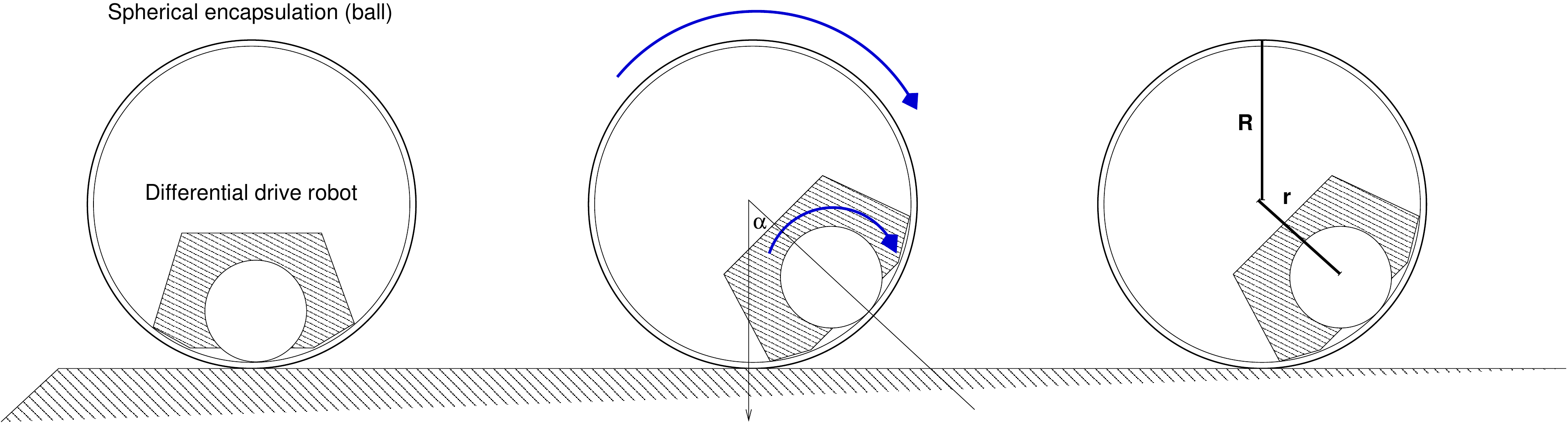}}
\centerline{\includegraphics[width=0.89\textwidth]{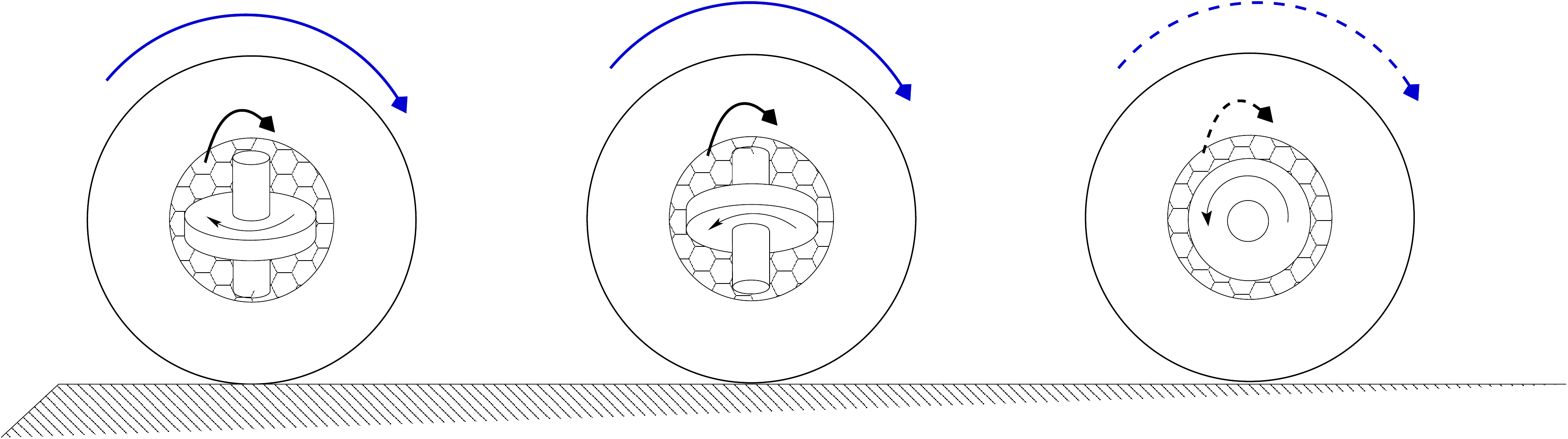}}
\caption{
\label{princ_fig} Different designs of encapsulated robots. Top row: typical design of robots like Sphero or Star Wars BB-8 toys. A weight is shifted along the inside of the hull which produces the torque to let the robot roll forward. Except for a very short transient period the torque is limited by the maximum angle $\alpha$ between the vertical direction and the inclination of the weight anisotropy inside the robot (theoretical limit is $90^o$). Bottom row: our design. Here the transient period can be extended considerably and the rotating gyro serves as a momentum reservoir until the rotation axis of the gyro is parallel to the rolling direction of the robot. The blue arrow indicates the direction of the torque.   
}

\end{figure*}

Comprehensive, encapsulated robots usually are formed into spheres (Sphero\footnote{https://www.sphero.com/}, Star Wars\footnote{Trademark of Lucasfilm Ltd., San Francisco, California} BB-8 toys, etc.), and they are actuated by a heavy weight that shifts along the inside of the thin wall of the robot's hull. This design allows for a limited torque and has other disadvantages versus the design that is presented here. Although those products from outside appear to work in a similar way the fundamental locomotion principle is quite different. One important advantage of our approach (see Fig. \ref{fig:robo_foto})
is that batteries and electronic units can be directly connected to the hull, seemingly rather a difficult task in the case of the toys on the market. 
In addition, in our approach the hull can have a much larger diameter than the actuator. An extended discussion of the advantages of the present approach can be found as part of the next section and as part of the discussion at the end of the paper.

Finally, T. Urakuboi introduced a new prototype of a spherical rolling robot with a gyro \cite{6284376,1631695}: It changes the direction of the gyro by driving the outer shell with two DC motors. The DC motor installed in the inner mechanism make a spherical outer shell rotate. By using the two motors, the angular momentum of the gyro is transferred to the outer shell and the robot rolls and moves on a surface. The gravity center of the robot is located at the geometric center of the sphere. The gyro rotates at a large angular velocity in the inside of the robot and has a large angular momentum. Different from these approaches, our design uses a structure that allows a reliable actuation system, where the motors are mounted on the shell structure. The actuation can be executed by driving a gear set. Thus, a comprehensive shelled robot is described here. 

In summary, our robot is moved by inertial torque actuation. To change direction by the rotate gimbal that represents a new type of idea for moving a robot. The structure of the remainder of this paper is: a section that explains the principle features of the new concept, the implementation and design details of our two robot prototypes, a brief overview over first experiments with the new robots, and finally a discussion, that also includes an outlook on possible applications. 

\begin{figure*}[!t]
\centerline{
\includegraphics[width=0.99\textwidth]{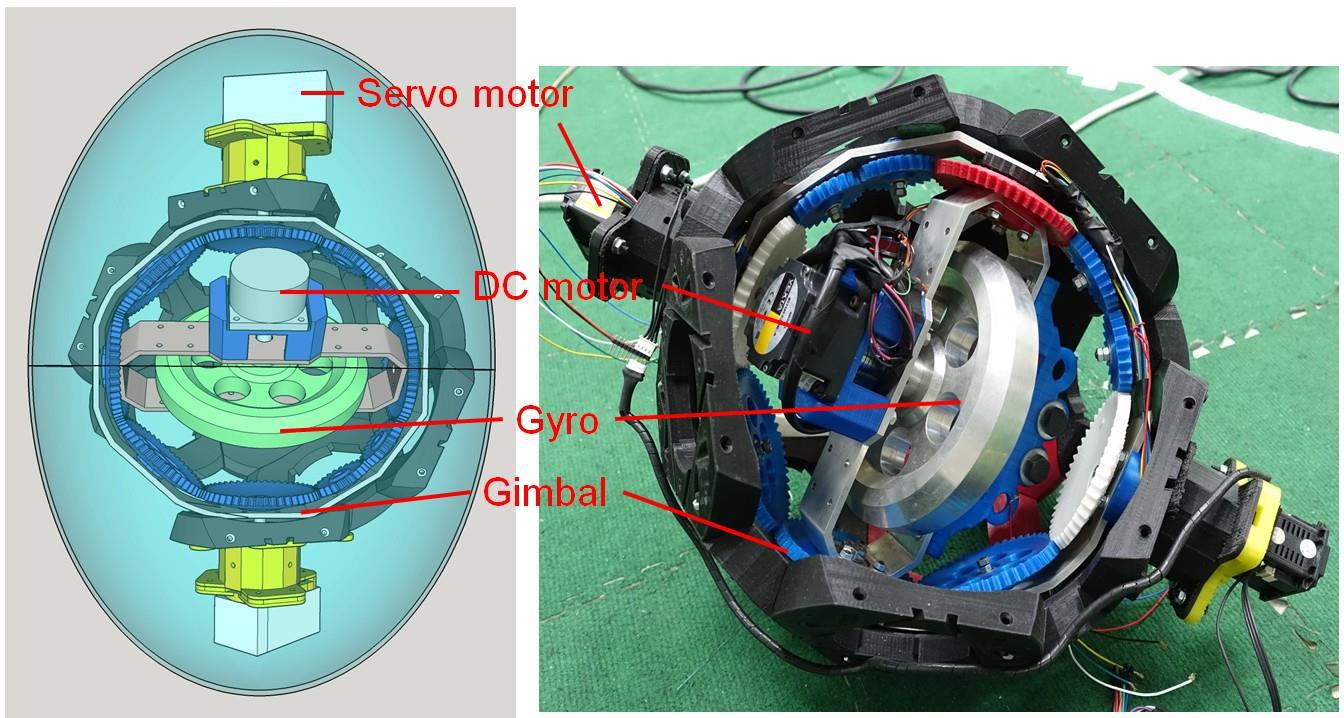}}
\caption{\label{fig:overview}
The overview of the mechanical architecture. Left: The actuation unit as it is located inside the egg shell. Right: The completely assembled actuator unit without the egg shell. The mechanical components are the gyro, a DC-motor that enforces the rotation of the gyro, a gimbal with two rings. The inside ring is designed to cover the gyro closely. The outside ring covers the inner gimbal, the DC motor, and the gyro. At the metal bend of the outside gyro we have put gears that can rotate freely. They allow to actuate the two rings of the gimbal. Finally, one can see two servo motors that actuate the two rings of the gimbal.
}
\end{figure*}

\section{Principle design features, advantages and limits}

In Fig. \ref{princ_fig} one can see the the principle design features of a class of standard encapsulated toy robots and our robot design. In off-the-shelf toy robots, a heavy weight is used to let the robot roll. One way to do this is to encapsulate a differential drive robot with two wheels inside a plastic ball (early Sphero type designs). Other types of designs use a hidden rotational axis inside the sphere where the weight can be rotated. The turning motion is then achieved by moving a weight on a sledge-like construction. In both cases the hull has to be very thin and it is relatively hard to implement functional parts in the outside structure (cameras or other types of sensors), because they interfere with the actuator as a necessary design feature. The maximal torque that can be used to drive the robot is limited by the mechanical design, which also limits the maximal angle of slopes that can be managed to climb.   

\begin{figure}[!b]
\centerline{\includegraphics[width=3.5in]{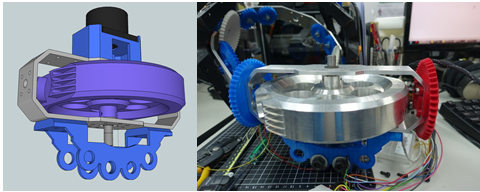}}
\caption{
\label{fig:inner_gimbal}
Assembled inner gimbal unit including gyro and DC motor. Design sketch shows the plastic parts (left). The screw-holes are used to attach the disk washers to balance out the mass of the DC-motor. Final part for the large prototype is shown on the right side.
}
\end{figure}

The key idea of the robot design presented here is that the big gyro in the middle of the robot serves as a reservoir (in the sense of a battery that supplies torque) of momentum from which torque can be elicited to drive the robot forward. The size of this reservoir is not restricted by the mechanical design directly, it is rather proportional to the speed of the rotating gyro. During operation the axis of the gyro shifts towards the rolling axis of the robot, i.e. the reservoir is exhausted. We consider two ways to restore the desired direction of the axis: The first is to use the friction between the robot and the ground to elicit momentum and the second to use the the shape of our robot and roll against the long side of the egg and in this way use the resulting gravitational momentum. A third option that is time and energy consuming is to stop the gyro, correct the axis and start over again. 

The design of the robot is arranged in a way that electronics, most motors, and even batteries can be located in the hull of the robot. All motions are actuated using gears. The actuation of the robot is realized by three motors, and only one of these three motors is used to keep the gyro at a constant rotation speed. Two of the motors are located at a fixed position versus the hull of the robot. They are used to move the axis of the gyro.  The features of the current design guarantee for a high reliability of the actuation compared to other designs, where most of the electric parts rotate versus the hull. 

Last but not least we used our custom 3D printer for most non-metal parts.


\begin{figure*}[!t]
\centering
\subfloat[Principle design]{\includegraphics[width=2.5in]{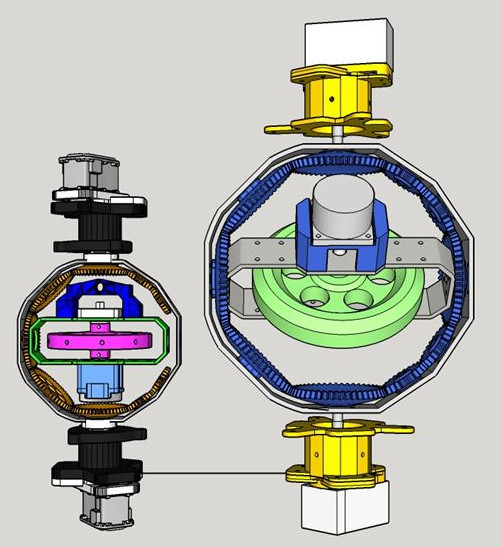}}
\label{subfig:ccc}
\hfil
\subfloat[S1  drives the gears clockwise and S2 anti-clockwise.]{\includegraphics[width=2in]{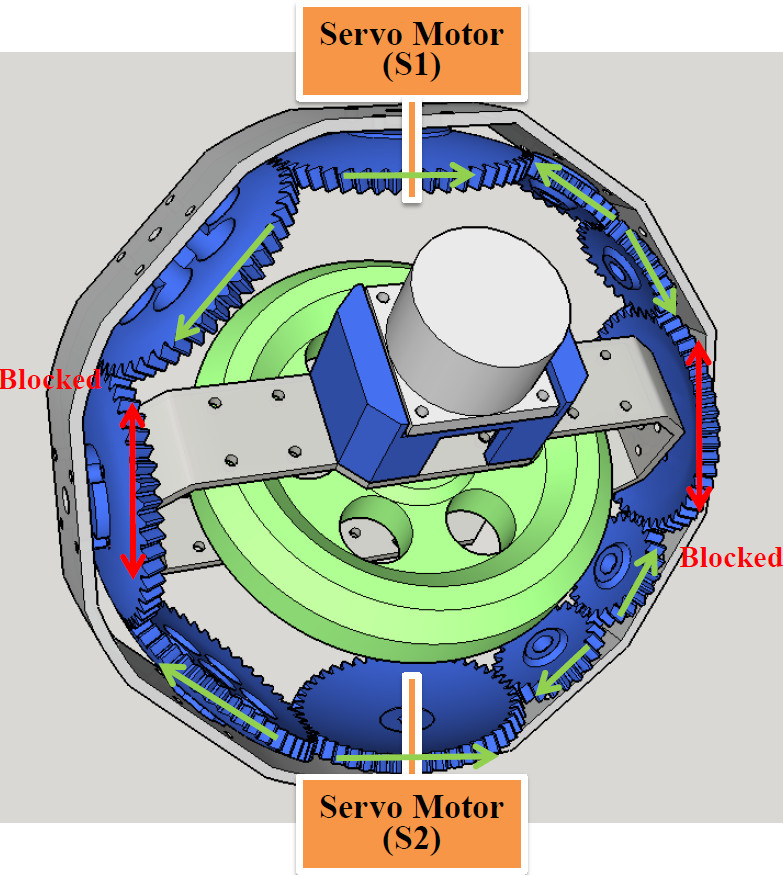}}
\label{subfig:cw and ccw}
\hfil
\subfloat[Both S1 and S2 rotate clockwise.]{\includegraphics[width=2in]{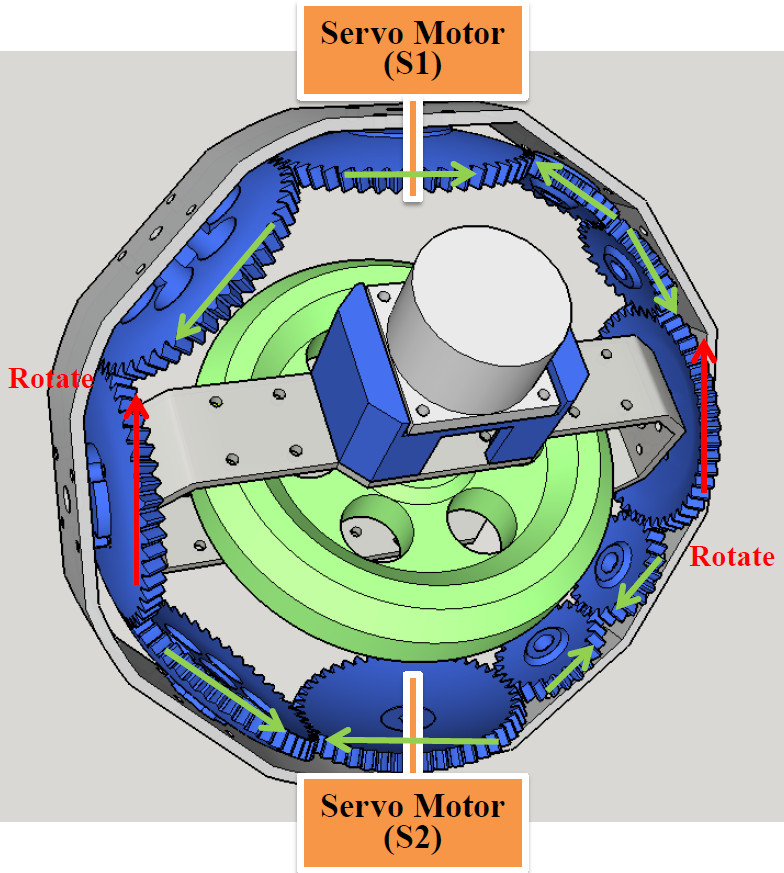}}
\label{subfig:cw}
\caption{
Mechanics and gear set to control the actuation of the gimbals. 
(a) Principle design of 1$^{st}$ (right) and 2$^{nd}$ prototype (left). Gears can be used in two different modes or by a linear superposition of them. In the first mode (b) the servo motors rotate in opposite direction (clockwise and anti-clockwise or vice versa) thus, the gears are blocked and that results in a rotation of the outer gimbal. In the second mode (c) the rotation of the gears transports the motion to the inner gimbal no force is elicited on the outer gimbal and it rests at the same position. Also, any linear superposition of both types can be applied.}\label{fig:The architecture of the gimbal set for both egg robots and rotation}

\end{figure*}

\section{Design Outline}\label{subsec:Design Concept of the Structure}

\subsection{Overview}

Fig. \ref{fig:overview} shows the entire inner mechanical structure of the egg robot. The gyro is located in the hollow sphere, which is in the center of the robot. The gyro consists of a rotor, an inner gimbal and an outer gimbal. The spherical parts of the robot are arranged into an icosahedron and form the inner shell. The inner shell was electrically connected to the gyro using slip rings. In this design, the inner shell is strong enough to endure the whole weight of the robot and can get sufficient space to taxi the gyro into any desired orientation so the gyro can rotate and change direction in this space. The battery and controller (not depicted) are deployed in the space between the inner shell and outer shell. 
			
The rings of the gimbal cannot rotate freely. Instead, they are actuated. There are two servo motors placed along the main symmetry axis of the egg. Both of them are connected to gears that are attached to the outer gimbal. When the two servo motors both rotate in the same direction, the inner gimbal rotates. If the two servo motors rotate in the opposite directions, the outer gimbal rotates. Thus, the gyro can be controlled to move in any direction through the servo motors.
			
Inside the robot is the heavy rotor. If the direction of the gyro is changed, the mechanic inertia changes the attitude of the robot and results in a force that affects the robot's movement. In this way, the robot can roll, rotate and spin.

\subsection{Gimbal design}			

The keystone of our design is the utilizatin of the actuated gimbal set. As mentioned before one feature of the design is to locate the controlling motors to the outside structure ("the hull") of the robot. Since the DC motor is inside the gimbal it is still necessary to have electrical connections through the rotation axes of the gimbal. This has been realized by using two slip rings that provide a multitude of electrical connections through these joints. Alongside these function specifications we have designed the gimbal structure. 

The inner gimbal (cf. Fig. \ref{fig:inner_gimbal}) is the basic frame for the rotor and the DC motor. It is a rigid metal part (6061 aluminium alloy) and thus provides a stable frame for a controlled motion system for the gyro; it also lets the DC motor work stably in high speed. Power lines and the signal lines are connected into the inner gimbal by means of 12 way slip rings. Slip rings are installed on one side of the inner gimbal. For balancing the weight of the DC motor, there are compensating disk washers at the opposite direction on the inner gimbal. Removing those disk washers would add an additional feature to the dynamics.  The outer gimbal is a pivoted support that allows the pitch rotation of the rotor and installs a gear set on it. It is made from 6061 aluminum alloy.

\begin{figure*} [!t]
\centerline{
\includegraphics[height=2.6in]{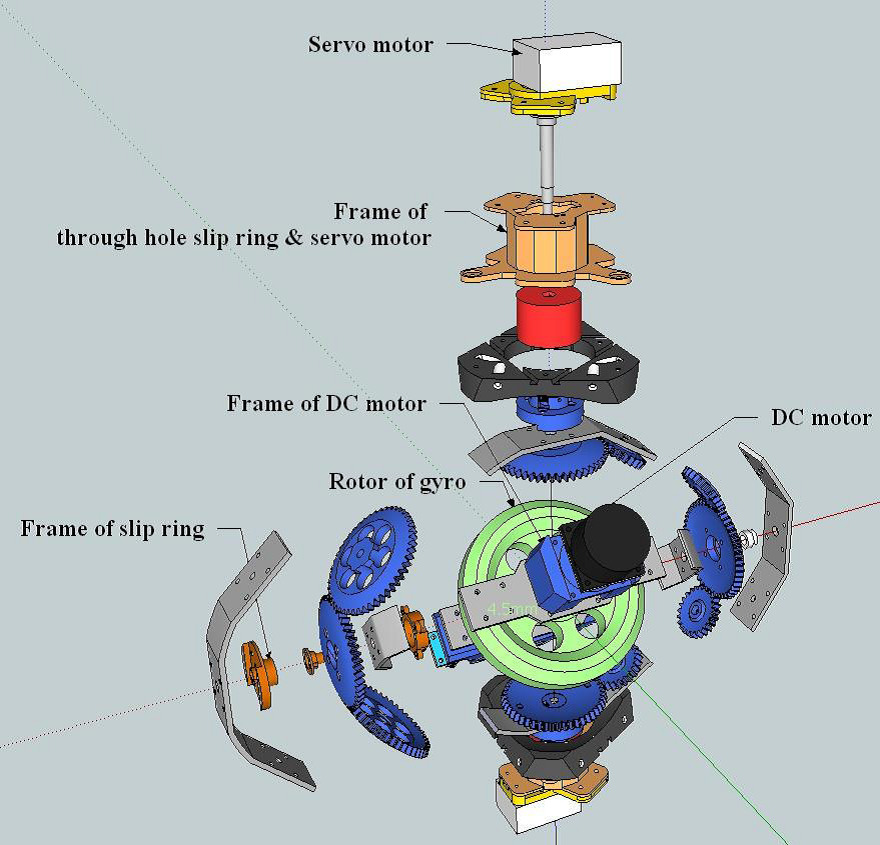} \hskip .3cm
\includegraphics[height=2.6in]{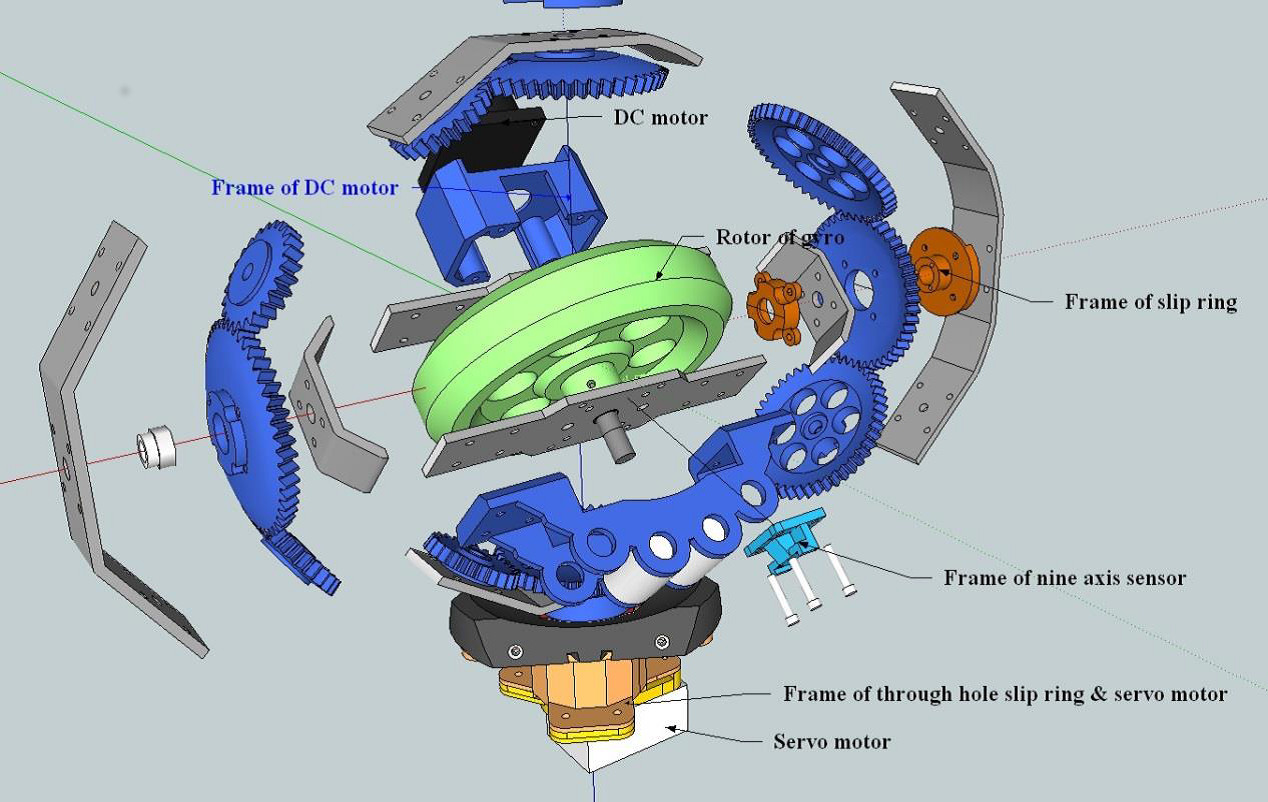} 
}
\caption{\label{explosion} The design concept of the inner structure for the robots, front view (left) and side view (right).}
\end{figure*}

Fig. \ref{fig:The architecture of the gimbal set for both egg robots and rotation} (a) shows the architecture of the outside ring of the gimbal set for both egg robots. The axis of both servo motors connects to the gears on the outer gimbal. Thus the servo motors can control the behavior of the gimbal by the gear set. When S1 rotates clockwise and S2 rotates anti-clockwise, or S1 rotates anti-clockwise and S2 rotates clockwise, the gears are blocked and the outer gimbal rotates (see Fig. \ref{fig:The architecture of the gimbal set for both egg robots and rotation} (b)). When S1 and S2 both rotate in a clockwise or anti-clockwise direction, the gears rotate. Thus, the inner gimbal rotates (see Fig. \ref{fig:The architecture of the gimbal set for both egg robots and rotation} (c)). In this way, the direction of the gyro can be changed. The coexistence of bearings, slip rings, gimbal rings and others requires a sophisticated mechanical design, which is illustrated in the explosion sketch of Fig. \ref{explosion}. 

\begin{figure*}[!t]
\centerline{
\includegraphics[height=2.8in]{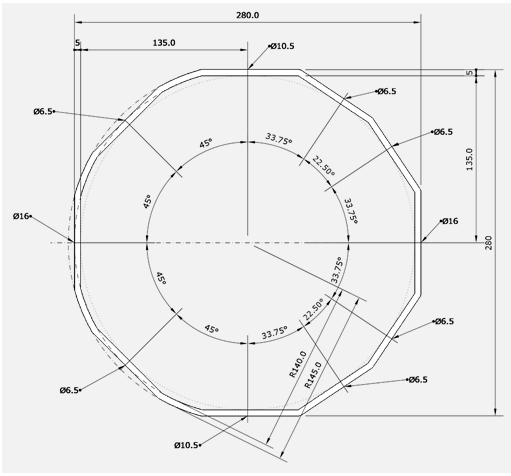} \hskip 2cm
\includegraphics[height=2.8in]{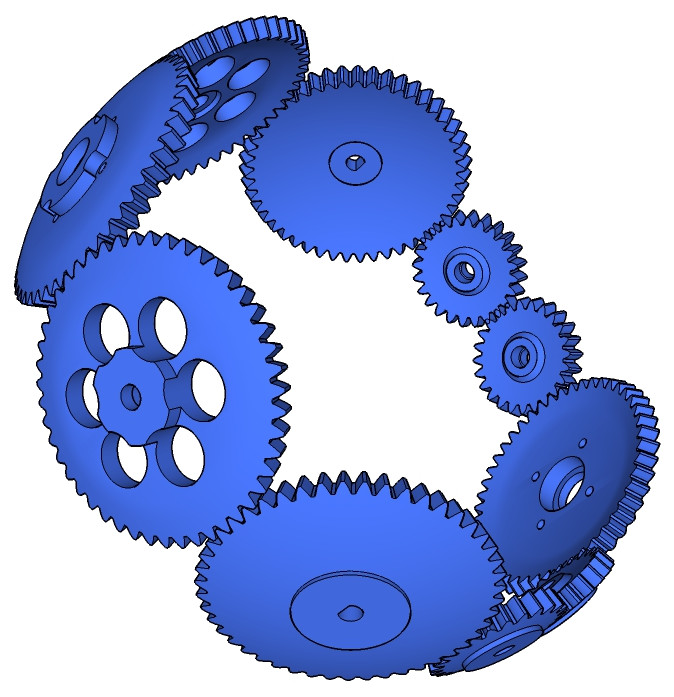}
}
\caption{\label{fig:gears} Outer gimbal ring design (left) and the gear set that is attached to the inside of the outer gimbal. One can see that the outer gimbal is not a ring. It rather forms an irregular polygon with design imprints to make space for each of the gears at their corresponding location.}
\end{figure*}

The transmission can be described by following equations.
				
\begin{equation}\label{equ:gimbal transmission}
\begin{bmatrix}\alpha \\ \beta\end{bmatrix} = \frac{1}{2} \begin{bmatrix}1 & 1 \\ 1 & -1\end{bmatrix}
\begin{bmatrix}\gamma_{S1} \\ \gamma_{S2}\end{bmatrix},
\end{equation}
where $\alpha$ is the angle of the outer gimbal, $\beta$ is the inner angle, and $\gamma_{S1}$ and $\gamma_{S2}$ are the angles of each of the servo motors. A detailed analysis of the kinematics and dynamics of the robot can be found in \cite{dynamics2018}.	

Ten gears are working in a series to transmit the torque in the mechanical system. 
All gears can rotate freely with respect to the outer gimbal ring (their frame). Two gears' axes are fixed to either servo motor, S1 and S2. Two more gears axes are connected to the inner gimbal and control its rotation angle.

There are two types of gears with different sizes, the ratio between a big gear and a small gear was intended to be roughly about 2:1.  The gears are not flat. Instead they are nestled closely to the shape of the outer gimbal. They have to be designed as bevel gears that are actuated in a circle. Thus, they are shaped as a cutout piece of a sphere. The four small gears placed at each side of the connection to inner gimbal are necessary to drive the inner gimbal on par and synchronously from both sides. 
				
In Fig. \ref{fig:gears} one can see the design of the outer gimbal ring and the composition of the gears. Spherical gears can be evaluated by using the theory of bevel gear sets where all axes are directed to the center of the gimbal ring, the curvature of the gears can be constructed as intersections to an imagined spherical shell. The diameter of the big gears is one side of an octagon. Similarly, the diameter of the small gear is one side of a hexadecagon. Both of two polygons are a tangential polygons that contain the circle which is a diameter of the thought circle of the gear set $r$. The diameter of each of the gears can then be calculated by the following equations.
				
\begin{equation}\label{equ:diameter of big gear}
D_{big-gear} = 2 r \tan(\frac{\pi}{8})
\end{equation}
and 				
\begin{equation}\label{equ:diameter of small gear}
D_{small-gear} = 2 r \tan(\frac{\pi}{16}).
\end{equation}
				
From the equations above, we calculated the ratio of both gears:
				
\begin{equation}\label{equ:diameter of gear ratio}
\frac{D_{small-gear}}{D_{big-gear}} = \frac{\tan(\frac{\pi}{16})}{\tan(\frac{\pi}{8})} = \frac{0.198912}{0.414213} = 0.480217
\end{equation}
				
Since the teeth sizes of all gears should be roughly equal and the ratio of gear teeth numbers should be coprime integers we chose the number ratio to be 48:23 (0.479167).

\subsection{Gyro and DC brushless motor}\label{subsec:DC brushless motor}
			
The DC brushless motor is used to drive the gyro. It is fixed on the inner gimbal. The type of the motor is BLH230K-A. The maximum speed is 3000 rpm. It includes break functionality, speed control, direction control, acceleration time setting and speed feedback. The BLH230K-A supports good stability with regard to its speed when there are load variations. The motor can compensate the moment and adjust to the goal speed immediately. In the current configuration, the gyro is used with the constant speed of 3000 rpm. Currently, breaking functionality is not being used. In the future it might add additional functionality to the gyro in the sense of a reaction wheel \cite{CLAWAR2004,jumpingjoe1,4058571}.
			
\subsection{Gyroscope and accelerometer}\label{subsec:Gyroscope and accelerometer}
The MPU-9255 is the Invensense’s first generation 9-axis activity detection part, working in conjunction with the AAR$^{TM}$ activity detection library. The digital output of this device can be I$^2$C or SPI. By reading the gyroscope sensor and accelerometer, the attitude of the robot can be detected while the rolling motion is occurring. We implemented two of these sensors on each robot. The first sensor is located at the hull, and the second sensor is located at the inner ring of the gimbal. 
				
\subsection{Controller}\label{subsec:Controller}
Due to the limited space in the egg robot, the main controller should be small and compact. We chose a credit card size computer named Raspberry Pi 2 to be the main controller. It uses an ARM Cortex-A7 CPU with 1 GB RAM. The clock rate is 900 MHz in medium mode but can be tuned to 1 GHz in overclocked mode. Thus, the computer can handle the computation for controlling the robot. The Raspberry Pi 2 contains extensive applications of standard interfaces, such as: USB host, HDMI output, Ethernet, UART, 
I$^2$C and GPIO (General-purpose input/output). We installed Raspbian, which is a free operating system based on Debian optimized for the Raspberry Pi, on the computer. Thus, the computer can be used as a low-cost controlling system with motor drivers and sensors on operating system-level or hardware-level development.

\begin{figure*}[!t]
\centerline{
\includegraphics[width=1.8in]{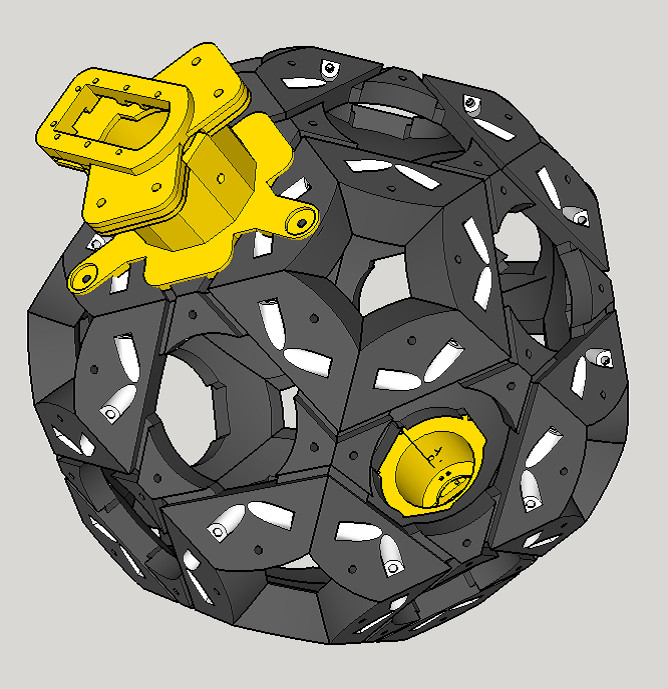} \hskip 1cm
\includegraphics[width=2.0in]{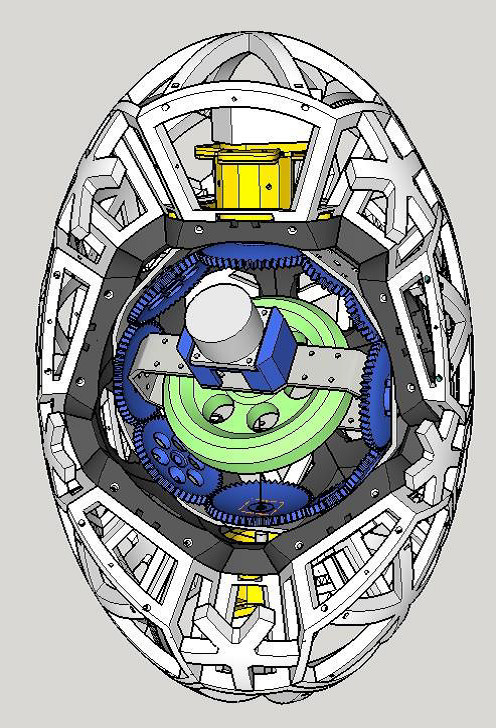} \hskip 1cm
\includegraphics[width=1.8in]{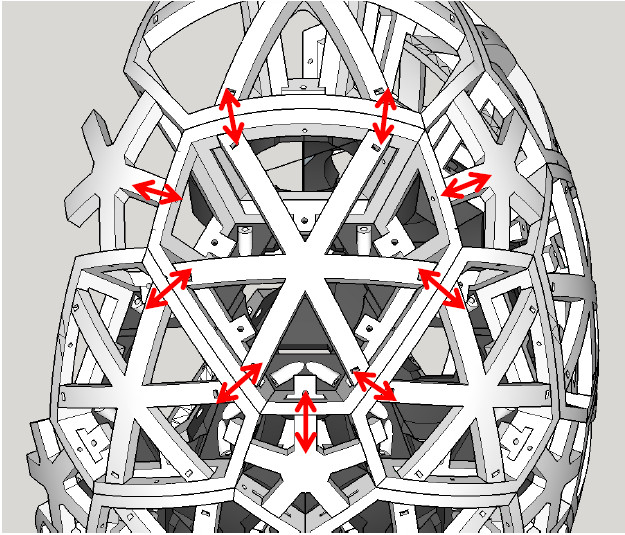}
}
\caption{\label{fig:shell} The inner shell (left) and outer shell (middle) are both based on truncated icosahedrons where in the case of the outer shell the icosahedron is elongated in one direction. Due to the bending of each tile, the pressure forces from outside are redirected to the neighboring tiles which guarantees by means of mutual tension an inert and sufficient stability of the robot.}
\end{figure*}

\begin{figure}[!b] \centerline{
\includegraphics[width=3.1in]{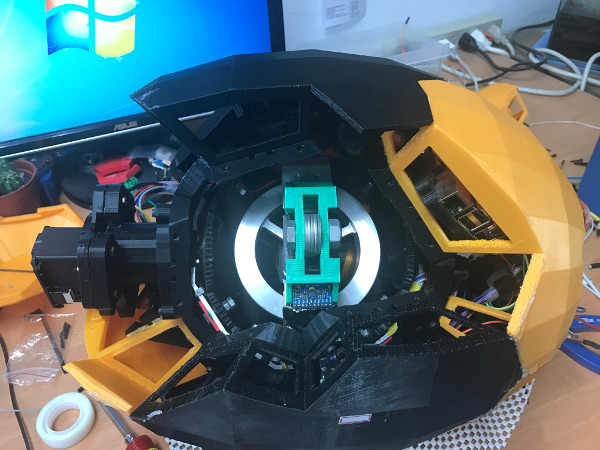}}
\caption{\label{fig:small_robot} Small robot prototype with open shell and view towards the gyro actuator.}
\end{figure}

\subsection{Servo motors}\label{subsec:servo motors}

The servo motors are used to drive the gimbal. The Dynamixel MX-106 and EX-106 are actuators, each one has a   speed reducer, magnetic encoder, controller, driver and network functionality. The servo motors use UART RS485 as a communication protocol for transferring data packages. The maximum baud rate is 1M bit per second. The data packages contain commands for the goal speed, goal position, goal torque and other controllable parameters. Also, we can get feedback by analyzing the packages from the servo motors. Thus, the controller has the ability to track the motor`s speed, position, temperature, voltage, current, and load.

\subsection{Structure and shell}\label{subsec:structure and shell}

The inner shell is constructed by a set of 20 plastic tiles as shown in Fig. \ref{fig:shell}. These tiles are organized in the shape of an icosahedron. The tiles had the shape of a regular triangle. In order to find a way around size restrictions of our rapid prototyping environment we truncated the edges of each triangle to be rounder. As a consequence, after all tiles were assembled, the sphere reached a sufficient size. In this way we also could create a space between the gimbal and outer shell. In order to make a stronger structure and smoother surface, we designed curved plates with six or five endpoints on the center of the pieces. Fig. \ref{fig:shell} at the left shows the pressure on the surface of the frame for each piece is transmitted to neighboring pieces. In this way, the outer shell is strong enough to support the whole weight of the robot.
				
Fig. \ref{fig:small_robot} shows the smaller egg robot during the assembling process. Due to its smaller size it was possible to merge tiles of the icosahedron for the prototyping process. So, it uses 14 pieces to build an ellipsoid outer shell. Each piece of the outer shell also stretches from the surface of the tiles of the inner shell. The design concept for this type is to make an ellipsoidal and sealed shell in order to build a water-proof and dust-proof structure that can roll smoothly. Each piece is a solid plate.

\begin{figure*}[!t]
\centerline{
\includegraphics[width=3.3in]{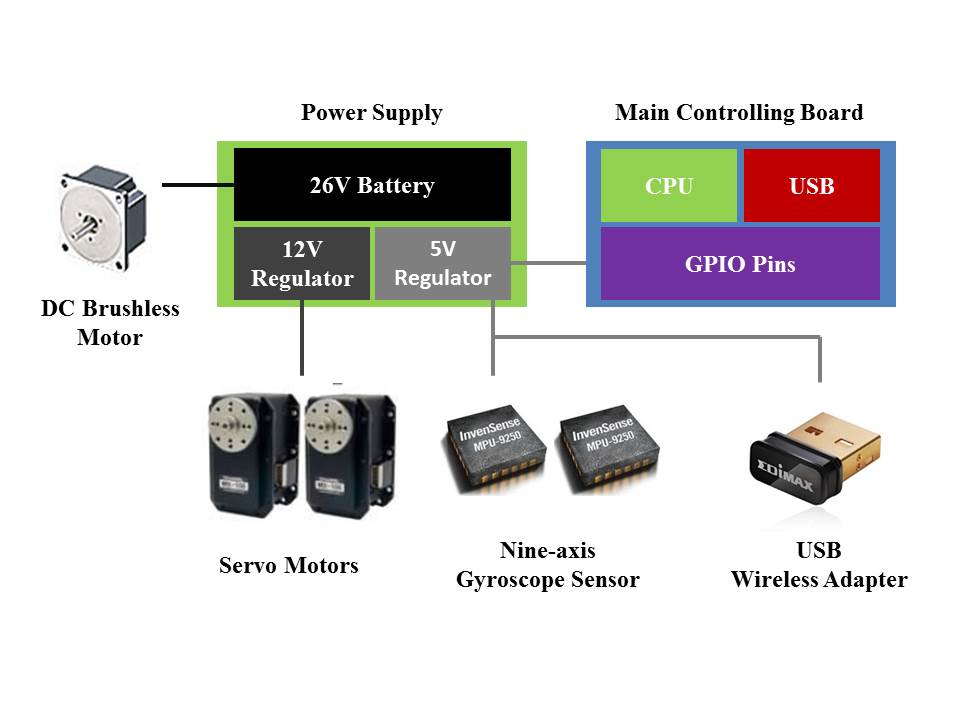} \hskip 1cm
\includegraphics[width=3.3in]{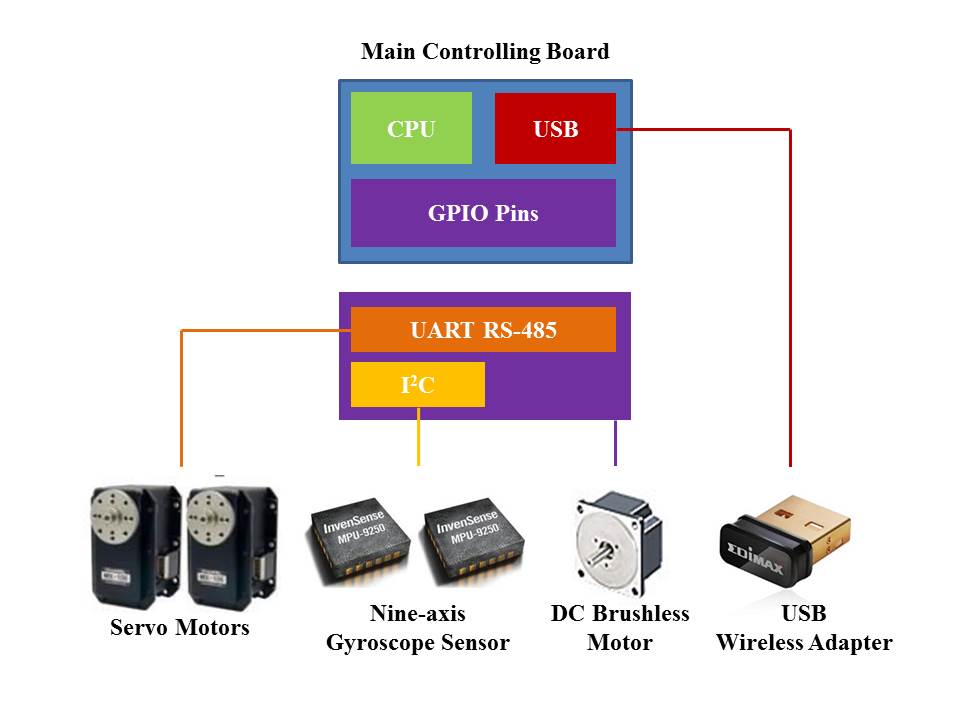} 
}
\caption{\label{fig:power} Power system (left) and control and communication system (right). }
\end{figure*}

\subsection{Power and system design}\label{subsec:power design}
			
A set of Sanyo 18650ZY 2600mAh 3.7V Li-on batteries is used as a power supply. Since the DC brushless motor requires a 24 V supply, seven of those batteries have been connected in series with an additional protection circuit. Fig. \ref{fig:power} shows the power sytstem in both prototypes. The circuit uses LM7812 IC for switching the input voltage to 12 V and LM2576S-5 for switching the input voltage to 5V. The two servomotors require a 12 V supply and at least 1.5 A for full power driving. This means we need two LM7812 ICs to supply 1 A of current. The control system and sensors need a total of about 5 V at 0.5 A. The voltage switching circuit can afford the load from the components. The runtime of our robots with a completely charged the battery set is about 1 hour if only the DC motor is enabled, and about 20~30 minutes if the robot is fully actuated during the entire time.

In our system design, we need a little embedded board to control all the motors and sensors. Fig. \ref{fig:power} (right) shows the frame of the control system. The Raspberry Pi 2 is used for the core board. It is used for processing data from sensors and controlling the motors and connecting to the outside world via WLAN. 
The two servo motors are connected to the GPIO and communicate with the board via the UART RS485 protocol. The nine-axes sensor which include gyro, accelerometer, magnetic and compass sensor is also connected by the GPIO. It uses I$^2$C to transmit the data. The DC brushless motor is controlled by its control board. With this board, we can control the speed, direction and motions of the DC motor. The board is connected to the GPIO pins. There is a USB adapter for the wireless networking board.

\begin{figure} [!b]
\centerline{
\includegraphics[height=1.7in]{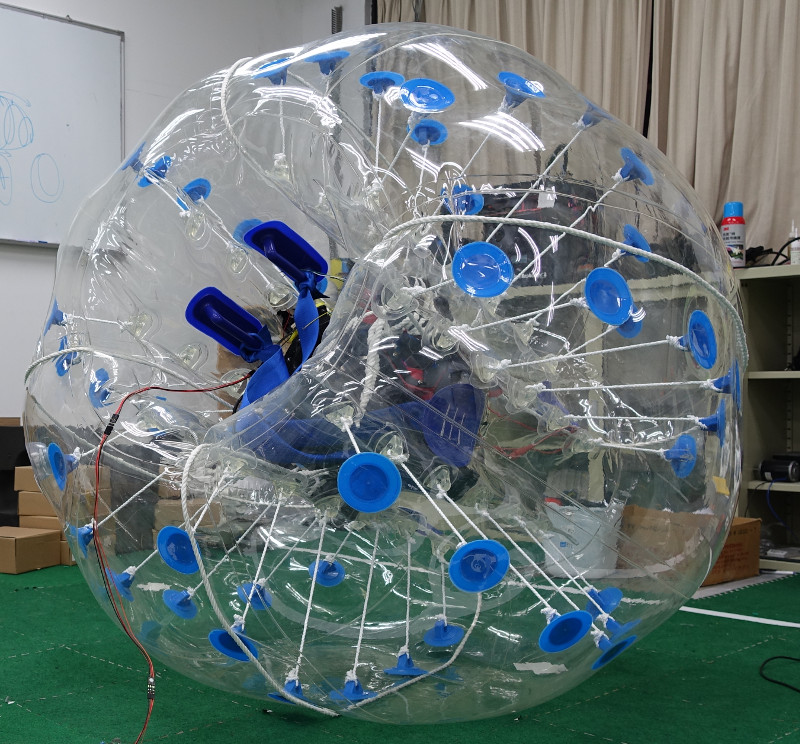} \hskip 0.3cm
\includegraphics[height=1.7in]{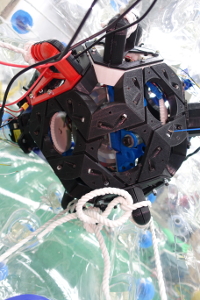} 
}
\caption{\label{ball} Gyro actuator core inside a toy ball (ca. 1.7 m diameter), complete setting (left) and gyro core inside the ball (right).}
\end{figure}

\section{Experiments}\label{sec:Experiment}

Both robots have been extensively tested and also have been used in demonstrations. The two prototypes work very reliably. The new prototype has also been used on water. Demonstrations can also be found in the multi-media attachment and in the internet\cite{bigegg2018}. In addition, the core has been used in experiments that used alternative shells. In one setting we put the core into a large balloon ball (cf. fig. \ref{ball}). The core could move the ball. Still, the ball was weakly actuated. However, we assume that if we increase the rotation speed from 3000 rpm to a significantly higher number the core is capable to control the ball. This setting can be seen as a prototype of a robot that is very large, but due to its light weight and soft cushion it is safe in the direct environment of pedestrians. It might also be combined with other types of locomotion that are currently tested in our laboratory, such as a robot that uses inflatable cushions \cite{ARIS2017}.

\section{Discussion and outlook}\label{sec:Conclusion and future work}

In this paper, we present a robot with no mechanism that directly interacts with the environment; it has a gyro in its center. Experiments with these robots show that using the angular momentum of the high speed rotating heavy rotor can work against the moment of inertia of the egg robot and make it move in a controlled fashion. Due to the enclosed shell, our robot prototypes and in a general sense our new type of actuator may be used in technical applications in various fields among rough environments such as underwater for work on underwater cables, sand, underground passage and even space and extraterrestrial environments, for example Venus with its high pressure and acidic atmosphere. The actuator core can work in all those environments by using the corresponding outer shell. In particular, we consider the following settings.
\begin{itemize}
\item Medical applications: A tiny version of the actuator core can be implemented into a robot pill that can be combined with sensors for use in several types of medical examinations, where the attitude of the pill can be conrolled.
\item Service and surveillance robots: The first studies into this direction have been outlined in Sect. \ref{sec:Experiment}. The robots can be large and still very light in weight and soft. These types of robots can be used in direct interaction with humans and used for advertising, eye catching, surveillance in parks and other things.
\item Mobile access point for wireless services. 
\end{itemize}
Some of these ideas have been submitted as patent applications. 
	
The next step is to make the egg robot more controllable. The control system can be improved by finding suitable algorithms or adding different sensors for a specific environment. Also, the egg robot could possibly be applied to practical applications in our daily life. The first step into this direction has been to derive an exact mathematical model of its dynamics, of which results have been published in \cite{dynamics2018}. A demo video of the larger prototype can be found at \cite{bigegg2018}.

\section*{Acknowledgment}
\addcontentsline{toc}{section}{Acknowledgment}
This work was supported by the Advanced Institute of Manufacturing with High-tech Innovations (AIM-HI), Ministry of Science and Technology under the MOST 103-2221-E-194-039, and 102-2221-E-194-050-program, and our industrial collaboration partner 1A Robotics.
	
\bibliographystyle{IEEEtran}
\bibliography{paper}
	
\end{document}